\documentclass[conference]{IEEEtran}
\IEEEoverridecommandlockouts
\usepackage{easy-todo}
\usepackage{cite}
\usepackage{amsmath,amssymb,amsfonts}
\usepackage{algorithmic}
\usepackage{graphicx}
\usepackage{textcomp}
\usepackage{xcolor}
\def\BibTeX{{\rm B\kern-.05em{\sc i\kern-.025em b}\kern-.08em
    T\kern-.1667em\lower.7ex\hbox{E}\kern-.125emX}}
\usepackage{amsthm}

\usepackage[UKenglish]{babel}

\usepackage[colorlinks]{hyperref}
\usepackage[nameinlink]{cleveref}

\usepackage{pgfplots}
\usetikzlibrary{patterns}
\usetikzlibrary{external}

\newtheorem{theorem}{Theorem}
\newtheorem{lemma}{Lemma}
\newtheorem{corollary}{Corollary}
\newtheorem{remark}{Remark}

\begin{document}

\title{Mean-Squared Accuracy of Good-Turing Estimator
\thanks{Supported by the FNR grant C17/IS/11613923}
}

\author{\IEEEauthorblockN{
Maciej Skorski}
\IEEEauthorblockA{
\textit{University of Luxembourg}}
}

\maketitle

\begin{abstract}
The brilliant method due to Good and Turing allows for estimating objects not occurring in a sample. The problem, known under names "sample coverage" or "missing mass" goes back to their cryptographic work during WWII, but over years has found has many applications, including language modeling, inference in ecology and estimation of  distribution properties.


This work characterizes the maximal mean-squared error of the Good-Turing estimator, for any sample \emph{and} alphabet size.
\end{abstract}


\section{Introduction}

\subsection{Background}


Let $X_1,\ldots,X_n\sim^{IID} p$ be a sample from a distribution $(p_s)_{s
\in S}$ on a countable alphabet $S$, and $f_s = \#\{i: X_i = s\}$ be the empirical (observed) frequencies. The missing mass
\begin{align}
    M_0 = \sum_{s}p_s\mathbb{I}(f_s=0),
\end{align}
which quantifies how much of the population is not covered by the sample, is of interest to statistics~\cite{robbins1968estimating} and several applied disciplines such as 
ecology~\cite{shen2003predicting,chao2003nonparametric,chao2013entropy,chao2017seen}, quantitative linguistic~\cite{efron1976estimating,mcneil1973estimating,gale1995good}, archaeology~\cite{myrberg2015tale}
network design~\cite{budianu2003estimation,budianu2004good}, information theory~\cite{vu2007coverage,zhang2012entropy}, and
 bio-molecular modeling~\cite{mao2002poisson,koukos2014application}. The most popular estimator due to Good-Turing~\cite{good1953population} is given by:
\begin{align}\label{eq:good_turing}
    \widehat{M} = \frac{1}{n}\sum_s \mathbb{I}(f_s = 1).
\end{align}
In this paper, the focus is on the maximal mean-squared error
\begin{align}\label{eq:mse_def}
    MSE = \mathbf{E}\left[(\widehat{M}-M_0)^2\right],
\end{align}
under the constrained alphabet (upper-bounded support)
\begin{align}
\# S = m.
\end{align}

\subsection{Related Work}

No prior work has studied the MSE under alphabet constraints;
we thus review the closest (in spirit) results of~\cite{rajaraman2017minimax} and~\cite{acharya2018improved} obtained for the unconstrained case $m=+\infty$.

The work~\cite{rajaraman2017minimax} expressed the mean-squared error \eqref{eq:mse_def} in terms of \emph{occupancy numbers}.
More precisely, define
\begin{align}
N_k \triangleq \sum_{s}\mathbb{I}(f_s=k),
\end{align}
the number of elements observed exactly $k$ times. Then~\cite{rajaraman2017minimax}:
\begin{align}\label{eq:mse_old}
    MSE = 
    \frac{\mathbf{E}\left[\frac{2N_2}{n} + \frac{N_1}{n}\left(1-\frac{N_1}{n}\right)\right]}{n} + O(n^{-2}),
\end{align}
and moreover $\max_{(p_s)\in\mathbb{P}(S)}MSE = \Theta(n^{-1})$ when $\#S=+\infty$, where $\mathbb{P}(S)$ is the set of probability measures on $S$.

The work~\cite{acharya2018improved} made an attempt to improve upon~\cite{rajaraman2017minimax} and establish a sharp constant (still with no constraint on alphabets); building on the formulas from~\cite{rajaraman2017minimax} and some further simplifications, it suggests to apply the method of Lagrange multipliers to prove that the maximum is achieved by a uniform distribution; noticeably, the proof was not given\footnote{The proof is omitted in both proceedings and public version.}

Beyond the scope of our problem are works on consistent estimation of missing mass~\cite{ohannessian2012rare,mossel2015impossibility}, concentration~\cite{esty1982confidence,esty1983normal,mcallester2000convergence,mcallester2003concentration,berend2013concentration,ben2017concentration}, expectation under constraints~\cite{berend2012missing,berend2017expected}, applications to distribution estimation~\cite{orlitsky2003always,orlitsky2015competitive,falahatgar2017power,hao2019doubly,hao2020profile}, and others~\cite{ayed2019good,cohen2021non,cohen2017cardinality}.


\subsection{Our Contribution}

\begin{itemize}
    \item We determine the worst MSE given the sample size $n$ and the alphabet size $m$; the maximizer is a Dirac-Uniform mixture with \emph{phase transition} depending on the ratio $\frac{m}{n}$.
    
    Studying the alphabet constraint is natural, but also practically motivated (prior support bound); it well fits other research on missing mass under constraints~\cite{berend2012missing,berend2017expected}; finally, it demonstrates useful optimization techniques.
    \item For $m=+\infty$, we obtain a \emph{rigorous} proof of the result stated in~\cite{acharya2018improved}. The method of Lagrange multipliers cannot be applied that easily;
    the main issue is the unbounded dimension, another is that maximizers for finite dimensions are more complicated than uniform distributions.
    \item Python implementation with examples, on GitHub~\cite{mskorski_git}.
\end{itemize}

\section{Results}


\subsection{Convenient Mean-Squared Error of Good-Turing Estimator}

We give the formula for MSE which involves \emph{first moments of occupancy numbers}, rather than variances as in prior works~\cite{rajaraman2017minimax,acharya2018improved}. Our expression is thus simpler to analyze.

\begin{theorem}\label{thm:good_variance_bound}
For any distribution $(p_s)$ we have:
\begin{align}\label{eq:mse_new}
        MSE = 
        \frac{\frac{2\mathbf{E}[N_2]}{n} + \frac{\mathbf{E}[N_1]}{n}\left(1-\frac{\mathbf{E}[N_1]}{n}\right)}{n} + O(n^{-2}),
\end{align}
The constant in $O(n^{-2})$ is independent of $(p_s)$ and $n$.
\end{theorem}

\subsection{Exponential Approximation / Poissonization}

Moments of occupancy numbers are often approximated with \emph{Poisson-like} expressions
~\cite{chao1992estimating,gnedin2007notes,zhang2009asymptotic}. We develop such an approximation, and use later for constrained optimization.

\begin{theorem}\label{thm:good_variance_bound_exp}
For any distribution $(p_s)$ we have:
\begin{multline}\label{eq:mse_new_exp}
MSE 
= \frac{n\sum_s p_s^2\mathrm{e}^{-np_s} + \sum_s p_s\mathrm{e}^{-np_s}-(\sum_s p_s\mathrm{e}^{-np_s})^2}{n} \\+ O(n^{-2}).
\end{multline}
\end{theorem}

\subsection{Extreme Mean-Squared Error Behavior}

Using non-linear programming (beyond Lagrange multipliers) we characterize the maximal MSE with respect to $\#S$. 
\begin{theorem}\label{thm:optimum}
For any $n\geqslant 2$ and $m=\#S \geqslant 2$, consider
\begin{align}
\begin{aligned}
\max &&
   \alpha(w,c)=w(1+c)\mathrm{e}^{-c}-(w\mathrm{e}^{-c})^2 \\
\mathrm{s.t.} &&   0\leqslant w \leqslant 1,\quad 
 w \leqslant \frac{m}{n} c.
\end{aligned}
\end{align}
Let $\alpha$ be the optimal value, and $c,w$ the optimal solution. Then it holds (also for $m=+\infty$) that:
\begin{align}
\max_{(p_s)\in\mathbb{P}(S)} MSE = \frac{\alpha}{n} + O(n^{-2}),
\end{align}
and this value is realized when $(p_s)$ is the mixture of the distribution uniform on $ \max\{\lfloor w n /c-1 \rfloor,1\}$ elements and the Dirac mass,
with the weights respectively $w$ and $1-w$.
\end{theorem}

\begin{figure}[h]
\centering
\begin{tikzpicture}[scale=0.65]
    \begin{axis}[view={-140}{60},xlabel = {$c$}, ylabel = {$w$},zmax=0.608,  y dir = reverse,colorbar]
    \addplot3[
        surf,
        shader = faceted interp,
        samples=50,
        domain=0:5,y domain=0:1
] 
        {y*(x+1)*exp(-x) - y^2*exp(-2*x)};
    \addplot3[
    black,ultra thick,dotted,
    domain=0:5,
    samples = 50,
    samples y = 0,
    y domain=0:1
]
    (
    {x},
    {min(0.8*x,1)},
    {min(0.8*x,1)*(x+1)*exp(-x)-min(0.8*x,1)^2*exp(-2*x)}
    );
    \end{axis}
\end{tikzpicture}
\caption{The optimization landscape of the program in \Cref{thm:optimum}.}
\label{fig:2d_program}
\end{figure}
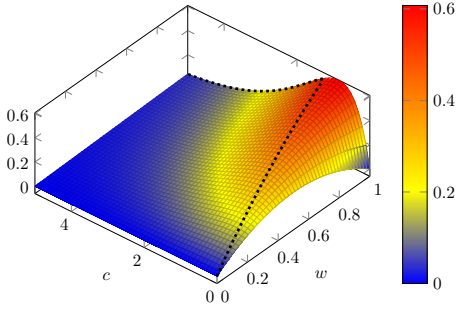

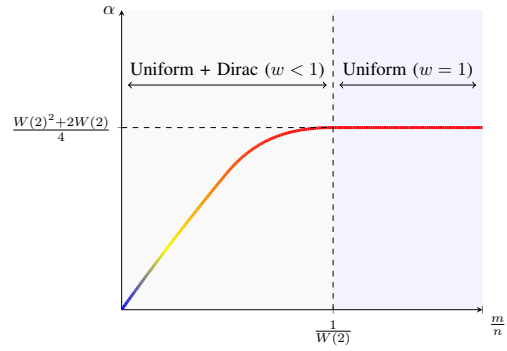
\begin{figure}[h]
\begin{tikzpicture}[scale=0.7]
\begin{axis}[
axis lines =middle,
xlabel = {$\frac{m}{n}$},
ylabel = {$\alpha$},
xtick={0, 1.173, 2},
xticklabels={0,$\frac{1}{W(2)}$},
yticklabels={0,$\frac{W(2)^2+2W(2)}{4}$},
ytick={0, 0.608},
ymax = 1,
every axis x label/.style={
    at={(ticklabel* cs:1.0)},
    anchor=north west,
},
every axis y label/.style={
    at={(ticklabel* cs:1.0)},
    anchor=east,
},
]
\addplot[mesh,ultra thick] table [x=b,y=mse,col sep=comma] {gt_risk.csv};
\addplot[mark=none,dashed,black] coordinates {(0,0.608) (1.172,0.608)};
\addplot[mark=none,dashed,black] coordinates {(1.172,0) (1.172,1.0)};
\path[fill=gray,fill opacity=0.05] (axis cs:0,0) -- (axis cs:0,1) -- (axis cs:1.172,1)--(axis cs:1.172,0);
\path[fill=blue,fill opacity=0.05] (axis cs:1.172,0) -- (axis cs:1.172,1) -- (axis cs:2,1.72)--(axis cs:2,0);
\node (n1) at (axis cs: 0,0.75) {};
\node (n2) at (axis cs: 1.172,0.75) {};
\node (n3) at (axis cs: 2.0,0.75) {};
\draw[<->] (n1)--(n2) node[midway,above] {Uniform + Dirac ($w<1$)};
\draw[<->] (n2)--(n3) node[midway,above] {Uniform ($w=1$) };
\end{axis}
\end{tikzpicture}
\caption{The leading constant in the max MSE (\Cref{cor:2d_solution}), depending on $m=\# S$. 
The phase transition (uniform maximizer) occurs at $m = \frac{n}{W(2)}$.}
\label{fig:phase}
\end{figure}

Below we show how to numerically compute the optimal value; $W(\cdot)$ denotes the Lambert-W function~\cite{corless1996lambertw}.
\begin{corollary}\label{cor:2d_solution}
Under the setup of \Cref{thm:optimum}:
\begin{align}
\alpha=\begin{cases}
\frac{W(2)^2+2W(2)}{4}= 0.608... & \frac{m}{n} \geqslant \frac{1}{W(2)} \\
\underset{0\leqslant c\leqslant \frac{n}{m}}{\max}\ \frac{mc(1+c)\mathrm{e}^{-c}}{n}-\left(\frac{mc\mathrm{e}^{-c}}{n}\right)^2 & \frac{m}{n}\leqslant \frac{1}{W(2)}.
\end{cases}
\end{align}
\end{corollary}
For illustration, see \Cref{fig:phase}; note the phase transition in the maximizer, not discussed in prior works. The specific "Uniform-Dirac" shape is also seen in other minimmax and optimization problems~\cite{berend2013concentration,bu2018estimation,obremski2017renyi,mohri2018foundations}.
Since
$\alpha=\Theta(\max\{\frac{m}{n},1\})$, the error $O(n^{-2})$ is negligible when $m\gg 1$.

\section{Preliminaries}    

In our estimations we need the following fact~\cite{johnson1994beta,gupta2004handbook}
\begin{lemma}[Mode of Beta-Distribution]\label{lemma:mode}
The expression $x^{a}(1-x)^{b}$ for $x\in [0,1]$, when $a,b>0$, is maximized at $x=\frac{a}{a+b}$.
\end{lemma}

Below we study in detail the function which particularly often comes up in our calculations (see also \Cref{fig:exp_quad}).
\begin{lemma}[Exponential-Quadratic Function]\label{lemma:exp_quad}
Define $g(u) \triangleq (u^2+bu)\mathrm{e}^{-u}$ for $u\geqslant 0$, with parameter $b\in\mathbb{R}$. Then : 
\begin{itemize}
\item $g$ has two local extremes $u = \frac{2-b\pm \sqrt{b^2+4}}{2}$ when $b<0$ and one extreme at $u = \frac{2-b+ \sqrt{b^2+4}}{2}$ when $b\geqslant 0$,
\item when $b<1$, $g$ is concave
in $\left(\frac{4-b- \sqrt{b^2+8}}{2},\frac{4-b+ \sqrt{b^2+8}}{2}\right)$ and
convex in $\left(0,\frac{4-b- \sqrt{b^2+8}}{2}\right)$ and $\left(\frac{4-b+ \sqrt{b^2+8}}{2},+\infty\right)$, 
\item when $b\geqslant 1$, $g$ is concave in $\left(0,\frac{4-b+ \sqrt{b^2+8}}{2}\right)$
and convex in $\left(\frac{4-b+ \sqrt{b^2+8}}{2},+\infty\right)$.
\end{itemize}
\end{lemma}
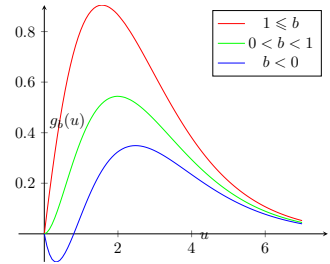
\begin{figure}[h]
\centering
\begin{tikzpicture}[scale=0.6]
\begin{axis}[
    axis lines =middle,
    xlabel = $u$,
    ylabel = {$g_b(u)$},
    enlarge x limits = true
]
\addplot [
    domain=0:7, 
    samples=100,
    red
]
{(x^2+1.2*x)*exp(-x)};
\addlegendentry{$1\leqslant b$}
\addplot [
    domain=0:7, 
    samples=100,
    green
]
{(x^2+0.01*x)*exp(-x)};
\addlegendentry{$0<b<1$}
\addplot [
    domain=0:7, 
    samples=100,
    blue
]
{(x^2-0.8*x)*exp(-x)};
\addlegendentry{$b<0$}
\end{axis}
\end{tikzpicture}
\caption{The auxiliary function $g_b(u) = (u^2+bu)\mathrm{e}^{-u}$, from \Cref{lemma:exp_quad}.}
\label{fig:exp_quad}
\end{figure}
Our proofs rely on non-linear optimization, particularly on the 
Karush–Kuhn–Tucker (first-order) conditions. For a detailed discussion we refer to optimization books~\cite{boyd2004convex,biegler2010nonlinear,bazaraa2013nonlinear}.
\begin{lemma}[KKT conditions]
Consider the program
\begin{align}
\begin{aligned}
\max && f(x) \\
\mathrm{s.t.} &&
\begin{cases}
h_i(x) = 0,& i\in I\\
g_j(x)\leqslant 0,& i\in J
\end{cases}
\end{aligned}
\end{align}
with differentiable real functions $f$,$(h_i)_{i\in I},(g_j)_{j\in J}$ in variables $x=x_1,\ldots,x_d$.
If the maximum occurs at $x$, then:
\begin{align}
\frac{\partial}{\partial x}f(x) = \sum_{i} \lambda_i \frac{\partial}{\partial x}h_i(x) +\sum_{j} \mu_j \frac{\partial}{\partial x}g_j(x)
\end{align}
where $\frac{\partial}{\partial x} = (\frac{\partial}{\partial x_1}\ldots\frac{\partial}{\partial x_d})$,
for $\lambda_i\in\mathbb{R}$, $\mu_j\geqslant 0$ such that
\begin{align}
\lambda_i \in\mathbb{R},\ \mu_j\geqslant 0,\ \mu_j g_j(x) = 0,
\end{align}
provided that regularity conditions hold at $x$.
\end{lemma}
We briefly remind the optimization terminology. The function $f$ is called objective, and any $x\in\mathbb{R}^d$ satisfying the constraints is called feasible. The optimal value is also called the program value. The constraint $h_i$ respectively $g_j$ is called active at $x$, when $h_i(x)=0$, respectively $g_j(x)=0$.

\begin{remark}[LICQ Constraints Qualification]\label{rem:licq}
The KKT conditions hold for optimal $x$ when the gradients of the constraints active at $x$ are linearly independent.
\end{remark}


\section{Proofs}

\subsection{Proof of \Cref{thm:good_variance_bound}}
Denote $\xi_s = \mathbb{I}(f_s=1)$. We have
\begin{align}
    \sum_{s\not=s'}\mathbf{E}[\xi_{s}\xi_{s'}] = n(n-1)\sum_{s\not=s'}p_{s}p_{s'}(1-p_{s}-p_{s'})^{n-2}.
\end{align}
Our goal is to estimate this expression up to $O(n)$.

We can assume that $p_s \leqslant \frac{1}{3}$. Indeed, since $(p_s)$ is a probability distribution, there are at most two values of $s$ such that $p_s>\frac{1}{3}$. The total contribution from all such $s$ to the right-hand side of the equation is at most $O(n^2 2^{-n}) \leqslant O(1)$.

We use the following bound, valid for $x\in [0,1]$ and $n\geqslant 1$:
\begin{align}
    (1-x)^{n} = 1-O(nx),
\end{align}
to $x=1-\frac{1-p_{s}-p_{s'}}{(1-p_{s})(1-p_{s'})} = \frac{p_{s}p_{s'}}{(1-p_{s})(1-p_{s'})} $, and
obtain:
\begin{align}
    \frac{(1-p_{s}-p_{s'})^{n-2}}{(1-p_{s})^{n-2}(1-p_{s'})^{n-2}} = 1-O(np_{s}p_{s'}).
\end{align}
This implies:
\begin{multline}
    \sum_{s\not=s'}\mathbf{E}[\xi_{s}\xi_{s'}] = 
     n(n-1)\sum_{s\not=s'}p_{s}p_{s'}(1-p_{s})^{n-2}(1-p_{s'})^{n-2} \\ +O(n),
\end{multline}
where we used $O(n\cdot n(n-1))\sum_{s\not=s'}p_{s}^2p_{s'}^2(1-p_{s})^{n-2}(1-p_{s'})^{n-2}= O(n^3)(\sum_{s}p_{s}^2(1-p_{s})^{n-2})^2=O(n)$; 
the last step follows by \Cref{lemma:mode} with $a=1, b=n-2$, and $\sum_s p_s=1$.

Furthermore, we have $\sum_{s=s'}p_{s}p_{s'}(1-p_{s})^{n-2}(1-p_{s'})^{n-2}=O(n^{-1})$ by 
\Cref{lemma:mode} applied to $a=1, b=2\cdot(n-2)$  and the condition $\sum_s p_s=1$; thus, we obtain:
\begin{multline}
    \sum_{s\not=s'}p_{s}p_{s'}(1-p_{s})^{n-2}(1-p_{s'})^{n-2}=\\=
    \left(\sum_{s}p_{s}(1-p_s)^{n-2}\right)^2 + O(n^{-1}).
\end{multline}
Using the last bound we conclude that:
\begin{align}
\sum_{s\not=s'}\mathbf{E}[\xi_{s}\xi_{s'}] = n^2\left(\sum_s p_s(1-p_s)^{n-2} \right)^2 + O(n).
\end{align}
In terms of the occupancy numbers $N_k$, we have shown that:
\begin{align}
    \mathbf{E}[N_1^2-N_1] = n^2\left(\sum_s p_s(1-p_s)^{n-2}\right)^2+O(n),
\end{align}
because $N_1 = \sum_{s}\xi_s$ and $\sum_{s\not=s'}\xi_{s}\xi_{s'} = N_1^2-N_1$. Finally,
$p_s(1-p_s)^{n-2} = p_s(1-p_s)^{n-1}\cdot(1+O(p_{s}))$ because 
$\frac{1}{1-p_s} = 1+O(p_s)$ for $p_s\leqslant \frac{1}{3}$; since $
\sum_s p_s^2(1-p_s)^{n-1} = O(n^{-1})$, by \Cref{lemma:mode} applied to $a=1,b=n-1$ and $\sum_s p_s=1$:
\begin{align}
\mathbf{E}[N_1^2-N_1] = n^2\left(\sum_s p_s(1-p_s)^{n-1}\right)^2+O(n),
\end{align}
and, since $n\sum_s p_s(1-p_s)^{n-1} = \mathbf{E}[N_1]$, we finally obtain:
\begin{align}
    \mathbf{E}[N_1^2-N_1] = \mathbf{E}[N_1]^2 + O(n).
\end{align}
Combining this with \Cref{eq:mse_old} and the fact that $\mathbf{E}N_1 = n\sum_{s}p_s(1-p_s)^{n-1})=O(n)$
finishes the proof.

\subsection{Proof of \Cref{thm:good_variance_bound_exp}}

The result can be derived from \Cref{thm:good_variance_bound} by relating Poisson and binomial distributions as in~\cite{barbour1984rate};  we give a direct argument. Since $\mathbf{E}[N_1] = n\sum_{s}p_s(1-p_{s})^{n-1}$
, $\mathbf{E}[N_2] = \binom{n}{2}\sum_{s}p_s^2(1-p_{s})^{n-2}$,
and $\sum_s p_s^2(1-p_s)^{n-2} = O(n^{-1})$ by \Cref{lemma:mode} with $a=1,b=n-2$, it suffices to show that:
\begin{align}
    \begin{aligned}
    \sum_{s}p_s(1-p_{s})^{n-1} &= \sum_{s}p_s\mathrm{e}^{-np_s} + O(n^{-1}) \\
    \sum_{s}p_s^2(1-p_s)^{n-2} & = \sum_s p_s^2\mathrm{e}^{-np_s} + O(n^{-2}).
    \end{aligned}    
\end{align}
We can assume $p_s\leqslant \frac{1}{3}$ (justified  as in the proof of \Cref{thm:good_variance_bound}).

Note that $(1-p)^n = \mathrm{e}^{-np}(1-x)^n$, $x=1-\frac{1-p}{\mathrm{e}^{-p}}$; $\frac{1}{2}\leqslant \frac{x}{p^2}\leqslant 1$ implies $(1-x)^{n}=1-O(nx)=1-O(np^2)$, thus:
\begin{align}
    (1-p)^n = \mathrm{e}^{-np}(1 - O(np^2)).
\end{align}

Moreover, we have the series of bounds:
\begin{align}
\begin{aligned}
    \sum_{s} p_s^k\mathrm{e}^{-np_s} & = O(n^{1-k}), \quad k=2,3,4 
\end{aligned}
\end{align}
obtained by introducing $q_s = 1-\mathrm{e}^{-p_s}$ so that 
$p_s^3\mathrm{e}^{-np_s} = O(1)q_s^3(1-q_s)^{n}$, using \Cref{lemma:mode} with  $a=1,b=n$, $a=2,b=n$ or $a=3,b=n$, and $\sum_s q_s = O(\sum_s p_s)=O(1)$.

These bounds finally give us:
\begin{align}
\begin{aligned}
    \sum_s p_s(1-p_s)^{n-1} & =  \sum_{s}p_s\mathrm{e}^{-(n-1)p_s} + O(n^{-1})\\
    \sum_s p^2_s(1-p_s)^{n-2} & =  \sum_{s}p^2_s\mathrm{e}^{-(n-2)p_s} + O(n^{-2}).
\end{aligned}
\end{align}

It remains to notice that
$e^{-(n-1)p}=\mathrm{e}^{-np}\cdot\mathrm{e}^p =
\mathrm{e}^{-np}(1+O(p))$, and similarly
$e^{-(n-2)p} = \mathrm{e}^{-np}\mathrm{e}^{2p} =\mathrm{e}^{-np}(1+O(p))$.
This means that replacing $n-1$ and $n-2$ above by $n$ we make the error of respectively $O(\sum_s p_s^2\mathrm{e}^{-np_s})=O(n^{-1})$ and $O(\sum_s p_s^3\mathrm{e}^{-np_s})=O(n^{-2})$. This completes the proof.

\subsection{Proof of \Cref{thm:optimum}}

\subsubsection{Non-Linear Programming }
In view of \Cref{thm:good_variance_bound_exp}:
\begin{align}
\max_{(p_s)} MSE = \frac{\alpha^{}}{n} + O(n^{-2}),
\end{align}
where $\alpha^{}$ is the value of the following optimization program:
\begin{align}\label{eq:orig_program}
\begin{aligned}
    \mathrm{max} && n\sum_s p_s^2\mathrm{e}^{-np_s} + \sum_s p_s\mathrm{e}^{-np_s}-(\sum_s p_s\mathrm{e}^{-np_s})^2\\
    \text{s.t.} && \forall s\in S: p_s \geqslant 0,\, \text{and}\, \sum_s p_s = 1,
\end{aligned}
\end{align}
and the optimal point gives the probability distribution realizing the maximum. We assume that $\#S=m<+\infty$, then the optimal solution $p^{*}$ exists (by the extreme value theorem~\cite{lovric2007vector}); we discuss $m=+\infty$  at the end of the proof.


\subsubsection{First-Order Conditions}

The KKT condition gives:
\begin{align}
   \forall s \in S:\quad 
a\cdot\frac{\partial }{\partial p_s}[p_s^2\mathrm{e}^{-np_s}] + b\cdot \frac{\partial }{\partial p_s}[p_s\mathrm{e}^{-np_s}] = \lambda, 
\end{align}
with coefficients $a,b$ that do not change with $s\in S$:
\begin{align}
    a \triangleq n,\quad b \triangleq  1-2\left(\sum_{s\in S} p_s\mathrm{e}^{-np_s} \right).
\end{align}
We conclude that for the optimal solution the components of $(p_s)_{s\in S}$ take values in the set of solutions $v$ to the equation:
\begin{align}
\frac{\partial}{\partial v}\left[(a v^2+bv)\mathrm{e}^{-nv}\right]=\lambda.
\end{align}

\subsubsection{Optimum is 3-Mixture}
We now argue that the equation has at most 3 positive solutions in $v$.
To this end, let us introduce $u=v n$ and use $a=n$ to simplify the equation:
\begin{align}
\frac{\partial}{\partial u}\left[(u^2+b u)\mathrm{e}^{-u}\right]=\lambda.
\end{align}
By \Cref{lemma:exp_quad}, the left-hand side changes its monotonicity at most twice. Thus, the equation has at most three solutions,
and the optimal $p_S$ takes at most 3 distinct non-zero values.

\subsubsection{6-D Program for 3-Mixture}

By the previous step:
\begin{align}
    \{p^{*}_s:p^{*}_s>0\} = \left\{\frac{c^{*}_1}{n},\frac{c^{*}_2}{n},\frac{c^{*}_3}{n}\right\},
\end{align}
with $c^{*}_i$ not necessarily distinct. Furthermore, let
\begin{align}
    m_i^{*} = \#\left\{s:p^{*}_s = \frac{c^{*}_i}{n}\right\}.
\end{align}
Then our original program \eqref{eq:orig_program} is equivalent to:
\begin{align}\label{eq:6d_p}
\begin{aligned}
\max &\
    \frac{\sum_{i=1}^{3}(m_i c_i^2+m_ic_i)\mathrm{e}^{-c_i}}{n}-\left(\frac{\sum_{i=1}^{3}m_ic_i\mathrm{e}^{-c_i}}{n}\right)^2 \\
\text{s.t.} &\  
\begin{cases}
 0\leqslant c_i  \text{ and } 0\leqslant m_i \text{ and } m_i \in \mathbb{Z},\quad i=1,2,3 \\
  \sum_{i=1}^{3}c_i m_i = n\\
  \sum_{i=1}^{3} m_i \leqslant m,
\end{cases}
\end{aligned}
\end{align}
with the optimal solution $(c^{*}_i),(m^{*}_i)$.


\subsubsection{Step 6: 2-D Program for Continuous Relaxation}

In the previous program $m_i$ are integers; we consider the relaxation
\begin{align}
\begin{aligned}
\max &\
    \frac{\sum_{i\in I}(m_i c_i^2+m_ic_i)\mathrm{e}^{-c_i}}{n}-\left(\frac{\sum_{i\in I}m_ic_i\mathrm{e}^{-c_i}}{n}\right)^2 \\
\text{s.t.} &\
\begin{cases}
 0\leqslant c_i  \text{ and } 0\leqslant m_i, \quad i\in I \\
  \sum_{i\in I}c_i m_i \leqslant n\\
  \sum_{i\in I} m_i \leqslant m,
\end{cases}
\end{aligned}
\end{align}
where $I=\{1,2,3\}$. We first prove that the maximum is achieved (not obvious, as $c_i$ are not bounded). Indeed, 
if the maximum is achieved as a limit with $c_i\to +\infty$, then the contributions to the objective
$m_ic_i^2\mathrm{e}^{-c_i}$ and $m_ic_i\mathrm{e}^{-c_i}$ tend to zero regardless of the values of $m_i$, because $m_i$ is bounded; in the limit we obtain the same value as when setting $c_i=0$ and an arbitrary fixed value for $m_i$ (this preserves the constraints).

We next argue that the maximum occurs at a point such that $m_i c_i = 0$ for at least two indices $i$; 
this means that we can assume $|I|=1$ and simplify the program to  two variables.
Suppose that $(c_i),(m_i)$ is optimal and such that the number $\#\{i:m_i c_i \not =0\}$ is smallest possible. 
If $\#\{i:m_i c_i  \not=0\}\leqslant 1$ there is nothing to prove, thus we assume $\#\{i:m_i c_i \not =0\}\geqslant 2$.
We can assume $m_i c_i \not =0$ for $i=1,2$ (due to the symmetry). The LICQ holds, as the gradients of \emph{possibly} active constraints
\begin{align}
\begin{aligned}
    \frac{\partial}{\partial [(c_i), (m_i)]} [c_3] & = (0,0,1,0,0,0) \\ 
    \frac{\partial}{\partial [(c_i), (m_i)]} [m_3] & = (0,0,0,0,0,1) \\ 
    \frac{\partial}{\partial [(c_i), (m_i)]}\left[\sum_{i=1}^{3}c_i m_i\right] &= 
    (m_1,m_2,m_3,c_1,c_2,c_3) \\
    \frac{\partial}{\partial [(c_i), (m_i)]}\left[\sum_{i=1}^{3}m_i \right]&=(0,0,0,1,1,1),
\end{aligned}
\end{align}
are linearly independent (here we use $m_1,m_2\not=0$). The KKT condition shows that $u=c_1,c_2$ satisfy the system:
\begin{align}
\begin{aligned}
 (u^2+b u)\mathrm{e}^{-u} &= n \lambda u + n\mu \\
 \frac{\partial}{\partial u}\left[(u^2+b u)\mathrm{e}^{-u} \right] &=
 n\lambda ,
\end{aligned}
\end{align}
where $b=1-\frac{2}{n}\sum_{i=1}^{3} m_i c_i \mathrm{e}^{-c_i}$; the first equation is the condition for $m_i$ multiplied by $n$, and
the second equation is the condition for $c_i$ multiplied by $n$ and divided by $m_i$ (here we use again $m_1,m_2\not=0$). 
Equivalently $g(u)\triangleq (u^2+bu)\mathrm{e}^{-u}$
is tangent to the straight line $n\lambda u + n\mu$ at 
points $u=c_1,c_2$. We claim this is not possible, unless $c_1=c_2$ (see \Cref{fig:exp_quad}). Indeed
$c_1\not=c_2$ must be on the different sides of the stationary point (by the mean-value theorem~\cite{matkowski2011mean,boyer1959history}); to match the slopes they need to be in the two intervals where $g$ decreases (\Cref{lemma:exp_quad}), so necessarily $b<0$; the intercept is below zero for the first interval (with the start-point at $0$), and above zero for the second interval (with the end-point at $+\infty$)
In turn, $c_1=c_2$ reduces to $c_1=0$; to see that
we define $c'_1,c'_2,c'_3=0,c_1+c_2,c_3$ and $m'_1,m'_2,m'_3=0,m_1+m_2,m_3$,
then replace $c_i,m_i$ by $c'_i,m'_i$, preserving the objective value and constraints.
But $\#\{i:m_i c'_i \not= 0\} = \#\{i:m_i c_i \not= 0\}-1$. This proves our claim that $m_i c_i = 0$ for two indices $i$. 

Our relaxed program is equivalent to:
\begin{align}\label{eq:2d_program}
\begin{aligned}
\max &&
    \frac{(m_1 c_1^2+m_1c_1)\mathrm{e}^{-c_1}}{n}-\left(\frac{m_1c_1\mathrm{e}^{-c_1}}{n}\right)^2 \\
\text{s.t.} &&  
\begin{cases}
 0\leqslant c_1 \text{ and } 0\leqslant m_1\\
  c_1 m_1 \leqslant n\\
  m_1 \leqslant m.
\end{cases}
\end{aligned}
\end{align}
\subsubsection{Relaxation Gap is Small}

We argue that the last step (relaxation) changes the optimal value by at most $O(n^{-1})$.

To this end, suppose that $c_1,m_1$ is optimal to \eqref{eq:2d_program} and let $P =  \frac{(m_1 c_1^2+m_1c_1)\mathrm{e}^{-c_1}}{n}-\left(\frac{m_1c_1\mathrm{e}^{-c_1}}{n}\right)^2$
; it suffices to construct 
$(c'_i),(m'_i)$ feasible for the program \eqref{eq:6d_p} and such that 
$P'=\frac{\sum_{i=1}^{3}(m'_i {c'_i}^2+m'_ic'_i)\mathrm{e}^{-c'_i}}{n}-\left(\frac{\sum_{i=1}^{3}m'_ic'_i\mathrm{e}^{-c'_i}}{n}\right)^2 $
satisfies
\begin{align}
P-P'\leqslant O(n^{-1}),
\end{align}
because the optimal value of \eqref{eq:6d_p} is upper-bounded by $P$ (by relaxation) and lower-bounded by $P'$  (by feasibility).

The optimal value is clearly positive and thus $m_1,c_1>0$. Define $m'_1 = \max\{\lfloor m_1-1 \rfloor,1\}$, $m'_2=1,m'_3=0$, also 
$c'_1 = m_1 c_1/m'_1$, $c'_2=n-m_1 c_1,c'_3=0$. Note that $m'_i,c'_i\geqslant 0$ and $m'_i$ are integers; moreover
$\sum_{i}m'_i \leqslant \max\{ m_1,2\}\leqslant m$ and $\sum_i m'_i c'_i = n$. Thus, $(c'_i),(m'_i)$ is feasible for \eqref{eq:6d_p}. 

The bound on $P-P'$ trivially follows when $m_1\leqslant 2$ because for $m_i =O(1)$ and $m'_i = O(1)$ we have $P = O(n^{-1})$, $|P'|= O(n^{-1})$. We further assume that $m_1\geqslant 2$, which implies $m'_1 = \Theta(m_1)$ and $c'_1 = \Theta(c_1)$. To bound $P-P'$ we observe that:
\begin{multline}
P-P'= \frac{m_1c_1}{n}( (c_1+1)\mathrm{e}^{-c_1}-(c'_1+1)\mathrm{e}^{-c'_1})\\
 -\frac{c'_2(c'_2+1)\mathrm{e}^{-c'_2}}{n} \\
 -\left(\frac{m_1c_1\mathrm{e}^{-c_1}}{n}\right)^2+ \left(\frac{m_1c_1\mathrm{e}^{-c'_1} + c'_2\mathrm{e}^{-c'_2}}{n}\right)^2,
\end{multline}
(we used $m_1c_1 = m'_1c'_1$ and $m'_2=1,m'_3=0$) so it remains to show that these three terms are at most $O(n^{-1})$.

The first term is
$\frac{m_1c_1}{n}\cdot (h(c_1)-h(c'_1))$ with $h(u)\triangleq (1+u)\mathrm{e}^{-u}$. By the mean-value theorem
$\frac{m_1c_1}{n}\cdot |h(c_1)-h(c'_1)| = \frac{m_1 c_1}{n}\cdot O(c_1\mathrm{e}^{-\Theta(c_1)})\cdot |c'_1-c_1|$;
since $|c'_1-c_1| = O(c_1/m'_1)=O(c_1/m_1)$, we can upper-bound as $\frac{O(c_1^2\mathrm{e}^{-\Theta(c_1)})}{n} = O(n^{-1})$.  

The second term is negative can be ignored.

The third term equals $x^2-y^2 = (x-y)(x+y)$ with 
$x=\frac{m_1c_1\mathrm{e}^{-c'_1} + c'_2\mathrm{e}^{-c'_2}}{n}$
and $y=\frac{m_1c_1\mathrm{e}^{-c_1}}{n}$. We have $0\leqslant x+y\leqslant O(1)$ by the constraint $m'_1c'_1\leqslant n$.
In turn,
$x-y = \frac{m_1c_1(\mathrm{e}^{-c_1}-\mathrm{e}^{-c'_1})}{n}+O(n^{-1})$. By the mean-value theorem applied to
$h(u) = \mathrm{e}^{-u}$ we get
$x-y = \frac{m_1 c_1}{n}\cdot |c_1'-c_1|\cdot O(1)\mathrm{e}^{-\Theta(c_1)}$; since 
$|c'_1-c_1| = O(c_1/m_1)$ we get  $|x-y|=\frac{O(c_1\mathrm{e}^{-\Theta(c_1)})}{n} = O(n^{-1})$ and $x^2-y^2\leqslant O(n^{-1})$.

\subsubsection{Summing Up}
With $w = \frac{c_1 m_1}{n},c_1=c$ we write \eqref{eq:2d_program} as:
\begin{align}
\begin{aligned}
\max &&
   w(1+c)\mathrm{e}^{-c}-(w\mathrm{e}^{-c})^2 \\
\text{s.t.} &&  
\begin{cases}
 0\leqslant w \leqslant 1 \\
 w \leqslant \frac{m}{n} c.
\end{cases}
\end{aligned}
\end{align}
We proved that the gap w.r.t \eqref{eq:orig_program} is  $O(n^{-1})$, and thus 
for the optimal value $\alpha$ the worst-case MSE equals $\frac{\alpha}{n} + O(n^{-2})$. The probability distribution achieving this value can be constructed from optimal $w,c$ as explained in the previous step; namely on $m'_1 = \max\{\lfloor m_1-1 \rfloor,1\}$ elements the probability mass equals $\frac{c_1}{n} = \frac{w}{m_1}$, and
on other $m'_2=1$ elements the probability mass is $\frac{n-m_1 c}{n}=1-w$.
This completes the proof when $m<+\infty$.

\subsubsection{Unbounded Dimension}
The supremum $\alpha$ of \eqref{eq:orig_program} can be approached on a sequence of distributions $(p^{(k)}_s)$, $k=1,2,\ldots$.

For any fixed $\epsilon > 0$, let $(p_s)$ be any distribution such that the the objective value is at least $\alpha-\epsilon$.
Since the objective of \eqref{eq:orig_program} is continuous with respect to the total variation distance, by a mass-shifting argument 
we can find $(p'_s)$ with finite support $m'$ such that the objective is at least $\alpha-2\epsilon$.
Replace $(p'_s)$ with $(p''_s)$ which is optimal under the support constraint $m'$; the objective can only increase, thus it is still at least $\alpha-2\epsilon$. When $m> \frac{n}{W(2)}$ the objective in \Cref{thm:optimum} has its maximum at $w=1$ and $c=W(2)$,
 as we will see in \Cref{cor:2d_solution}, and then the constraint $w\leqslant \frac{m}{n}$ can be ignored.
This shows that limiting the support to $m=O(n)$ we can approximate the supremum up to $2\epsilon$, for arbitrarily small $\epsilon$. Thus, for $m=+\infty$ the theorem also holds (the constraint with $m=+\infty$ is automatically satisfied, hence to be ignored).

\subsection{Proof of \Cref{cor:2d_solution}}

We denote $b\triangleq\frac{m}{n}$ and consider the program:
\begin{align}
\begin{aligned}
\max &&
   w(1+c)\mathrm{e}^{-c}-(w\mathrm{e}^{-c})^2 \\
\text{s.t.} &&  
 0\leqslant w \leqslant 1 \ \textrm{ and }
 w \leqslant b c
\end{aligned}
\end{align}
with respect to the parameter $b>0$. 

The program has no local maximum with $0<w<1$ and $w<bc$ (also seen in \Cref{fig:2d_program}).
Indeed, otherwise the first order condition for $w$ would give 
$w=\frac{(1+c)\mathrm{e}^c}{2}$; the program becomes maximizing
$\frac{1+c}{2}-\left(\frac{1+c}{2}\right)^2$ subject to $\frac{(1+c)\mathrm{e}^c}{2}<b c$,
and the first order condition for $c$ gives $c=0$, a contradiction.

Since at the optimal point $w\not=0$ (the objective would be zero), we are left with two cases.
For $w=1$ the program is
\begin{align}\label{eq:2d_case1}
\max_{\frac{1}{b}\leqslant c}\ (1+c)\mathrm{e}^{-c}-\mathrm{e}^{-2c},
\end{align}
and when $w=bc$ the program becomes
\begin{align}\label{eq:2d_case2}
\max_{c\leqslant \frac{1}{b}}\  b(c+c^2)\mathrm{e}^{-c}-b^2c^2\mathrm{e}^{-2c}.
\end{align}
In what follows we use the fact that $h(c)\triangleq (1+c)\mathrm{e}^{-c}-\mathrm{e}^{-2c}$ increases for $0<c<W(2)$ and decreases when $W(2)<c$.

Suppose that $b\leqslant \frac{1}{W(2)}$; then \eqref{eq:2d_case1} is maximized at $c=\frac{1}{b}$;
this matches the objective of \eqref{eq:2d_case2} at $c=\frac{1}{b}$, so the optimal value of $\eqref{eq:2d_case2}$ is bigger or equal than that of \eqref{eq:2d_case1}.

Suppose now that $b\geqslant \frac{1}{W(2)}$, 
then \eqref{eq:2d_case1} is maximized at $c=W(2)$, with the value $h(W(2))$. This time we claim that \eqref{eq:2d_case2} is smaller or equal than  \eqref{eq:2d_case1}.  If $\frac{(1+c)\mathrm{e}^c}{2}<1$ for optimal $c$, then (under the constraint) we upper-bound the objective as $bc \cdot 2c \mathrm{e}^{-2c}\leqslant 2c\mathrm{e}^{-2c}\leqslant \frac{1}{\mathrm{e}}<0.608...=h(W(2))$. When $1 \leqslant \frac{(c+1)\mathrm{e}^c}{2}$ then, under the constraint, we have
$b\leqslant \frac{1}{c}\leqslant \frac{(c+1)\mathrm{e}^c}{2c}$, and 
since the objective increases in $b\leqslant \frac{(c+1)\mathrm{e}^c}{2c}$ setting $b=\frac{1}{c}$ gives the upper bound $h(c)$; this in turn at most $h(W(2))$.

\section{Conclusion}

This work determines the worst mean-squared error of the Good-Turing estimator given the sample and alphabet size, completing upon prior results for unrestricted distributions.








\bibliographystyle{./bibliography/IEEEtran}
\bibliography{citations}


\end{document}